\documentclass[conference]{IEEEtran}
\IEEEoverridecommandlockouts

\usepackage{cite}
\usepackage{amsmath,amssymb,amsfonts}
\usepackage{algorithmic}
\usepackage{graphicx}
\usepackage{textcomp}
\usepackage{xcolor}
\usepackage{multirow}
\usepackage{booktabs}
\usepackage{hyperref}
\usepackage[T1]{fontenc}
\usepackage{float}
\usepackage{array}
\usepackage[table]{xcolor} 

\def\BibTeX{{\rm B\kern-.05em{\sc i\kern-.025em b}\kern-.08em
    T\kern-.1667em\lower.7ex\hbox{E}\kern-.125emX}}
\begin{document}

\newcommand{\red}[1]{\textcolor{red}{#1}}
\newcommand{\graycell}[0]{\cellcolor{gray!15}}
\newcommand{\skycell}[0]{\cellcolor{blue!15}}

\title{Multi-Objective Instruction-Aware Representation Learning in Procedural Content Generation Reinforcement Learning}
\author{\IEEEauthorblockN{Sung-Hyun Kim, Geum-Hwan Hwang, In-Chang Baek, Seo-Young Lee, Kyung-Joong Kim}
\IEEEauthorblockA{
\textit{Gwangju Institute of Science and Technology (GIST), South Korea} \\
\{\texttt{st4889ha}\}{@gm.gist.ac.kr} \\
}
}

\maketitle
\begin{abstract}
Recent advancements in generative modeling have highlighted natural language as a highly expressive and accessible modality for controlling content generation.
However, existing instruction-conditioned reinforcement learning methods for procedural content generation often struggle to fully exploit the expressive richness of textual input, particularly under complex multi-objective instructions, resulting in limited controllability.
To address this limitation, we propose \textit{MIPCGRL}, a multi-objective representation learning method for instructed content generators that incorporates sentence embeddings as conditioning signals.
Experimental results show that the proposed method achieves a 13.2\% average improvement in controllability under multi-objective instructions.
The learned representation also generalizes to instructions specifying previously unseen subsets of objectives derived from training-time multi-objective compositions, while preserving objective-specific factors.
This ability to process complex instructions enables more expressive and flexible content generation.

\end{abstract}

\begin{IEEEkeywords}
procedural content generation, text-to-level generation, reinforcement learning, natural language processing, multi-objective learning
\end{IEEEkeywords}

\newcolumntype{C}[1]{>{\centering\arraybackslash}p{#1}}
\newcolumntype{L}[1]{>{\raggedright\arraybackslash}p{#1}}

\section{Introduction}

Recent advances in generative modeling have positioned natural language as a practical high-level interface for content creation, allowing users to express rich constraints through prompts rather than low-level parameters. Large-scale text-conditioned diffusion models have demonstrated strong controllability for visual synthesis~\cite{ho2020ddpm,ramesh2021zeroshot,rombach2022high,saharia2022photorealistic}, and prompt conditioning has been extended toward more complex generative settings such as text-to-video and temporally consistent generation~\cite{singer2022makeavideo,ho2022videodiffusion}. Similar prompt-driven trends appear in interactive media and games: language prompts can guide the generation of tile-based levels~\cite{sudhakaran2023mariogpt,todd2023level} and can condition lightweight generators to produce small-domain assets such as maps and sprites from sentence embeddings~\cite{merino2023fivedollar}. Collectively, these results suggest that natural language can serve not merely as a label, but as a compositional specification language for multi-constraint content generation.

\begin{figure}[!t]
    \centering
    \includegraphics[width=1.0\linewidth]{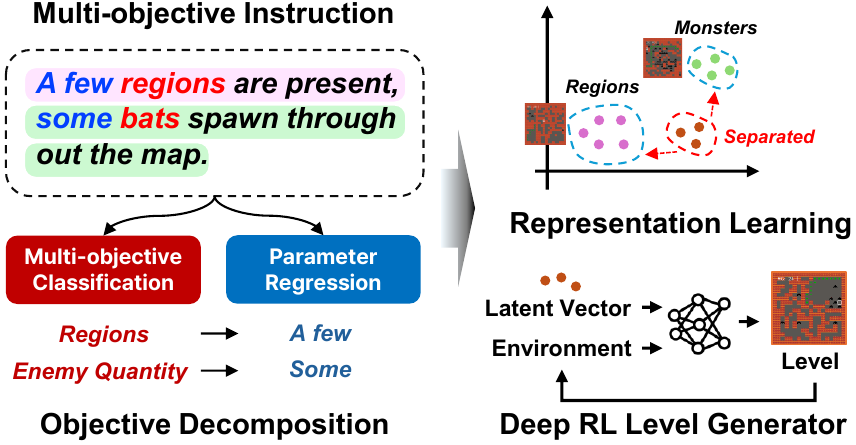}
    \vspace{-0.6cm}
    \caption{\textbf{MIPCGRL Overview.} We encode a compositional natural-language instruction into objective-aware latent factors (active-objective classifier + condition-regression head), which conditions an RL-based generator to produce levels aligned with the separated constraints.}
    \label{fig:teaser}
    \vspace{-15pt}
\end{figure}

Within game content generation, controllable procedural content generation via reinforcement learning (PCGRL) \cite{pcgrl} offers a compelling alternative to data-free generative pipelines: instead of requiring large curated datasets, it enables goal-directed generation by optimizing reward functions that encode desired content properties. However, traditional controllable PCGRL \cite{cpcgrl} commonly exposes its control interface through predefined scalar parameters, which constrains expressivity from a game-design perspective and reduces accessibility for broader user groups—effectively limiting how designers can externalize creative intent into the generator’s input space. To improve usability and expressiveness, IPCGRL \cite{ipcgrl} introduced natural-language control by learning a semantic latent space in which instructions condition an RL policy. Yet practical design requests are often compositional, involving multiple objectives within a single utterance, and IPCGRL struggles to represent such multi-objective conditions due to the limited expressive capacity of its simple text-encoder architecture. As a result, even though instruction conditioning is feasible for simpler instructions, the learned RL policy may fail to reliably generate content that satisfies multiple constraints simultaneously under compound commands.

To address this limitation, we propose \textbf{M}ulti-objective \textbf{I}nstruction \textbf{PCGRL} (MIPCGRL), which extends IPCGRL with a modular representation-learning architecture designed explicitly for multi-objective instruction semantics. 
Fig.~\ref{fig:teaser} illustrates our pipeline, where a compositional natural-language instruction is encoded into an objective-aware latent code that conditions the generator to produce levels satisfying the combined constraints.
The key idea is to reduce semantic interference by disentangling objective-specific representations and selectively activating them based on the instruction. Concretely, MIPCGRL introduces a multi-label objective classifier that predicts which predefined objectives are active in an input instruction and uses these probabilities to control latent-space activation. To support this, a multi-head fitness regression module decomposes the latent representation into objective-specific vectors. This enables probabilistic weighting to suppress irrelevant objective components while preserving components relevant to the instruction. The resulting weighted representation is then used as the conditioning signal for instruction-conditioned RL, enabling the policy to better interpret and execute compound design intents.

We evaluate whether MIPCGRL (i) improves performance on compositional multi-objective instructions while remaining robust under single-objective instructions; (ii) generalizes to subset compositions by remaining valid and consistent on instruction sets formed by selecting a subset of objectives from the multi-objective combinations used during training; and (iii) exhibits improved trade-off behavior by comparing regions of simultaneous improvement and objective conflicts through a density analysis of multi-objective progress.

This paper makes the following key contributions:

\begin{itemize}
\item We propose MIPCGRL, a structurally improved instruction representation framework for learning objective-aware representations in language-conditioned multi-objective PCGRL.
\item We show that MIPCGRL improves performance on multi-objective instructions while remaining robust under single-objective instructions.
\item We further analyze its behavior under more complex triplet instructions through subset-composition evaluations and ablation studies.
\end{itemize}

\section{Background}

\subsection{Language-Instructed PCGRL}

Controllable PCGRL \cite{cpcgrl} commonly conditions an RL generator on designer-specified targets, often expressed as scalar objectives or manually tuned parameters.
While effective for goal-directed generation, such interfaces typically require translating high-level design intent into low-level numeric specifications.
To make conditioning more designer-friendly, IPCGRL~\cite{ipcgrl} introduced natural language as a control modality by using a text embedding space as the conditioning feature for RL, treating the instruction as a generation goal.

Given a natural language instruction $\mathcal{I}$, IPCGRL maps it into a latent space in two steps: (1) a pretrained \emph{Bidirectional Encoder Representations from Transformers (BERT)} model~\cite{bert} embeds the token sequence into a general-purpose representation,
$z_{\text{bert}} = \text{BERT}_{_\psi}(\mathcal{I})$, and (2) an objective-specific encoder $E_{\theta}$ maps it to a domain-adapted embedding,
$z_{\text{enc}} = E_{\theta}(z_{\text{bert}})$, intended to capture game-relevant semantics.
During RL training, the embedding is concatenated with the environment state, $o_t = \{z_{\text{enc}}, s_t\}$, and fed into the policy network $\pi(a_t \mid o_t)$ to learn instruction-conditioned generation policies over a fixed instruction set.
The encoder in~\cite{ipcgrl} is trained on single-objective instructions using a lightweight regression architecture and an offline state buffer $\mathcal{B}$, which aligns the instruction embedding space with the environment state distribution and provides supervision for the regression objective.

However, IPCGRL’s encoder exhibits a structural limitation due to its \emph{single regression head}. 
Because instruction effects are predicted through a single scalar objective, the model lacks an explicit mechanism to disentangle (i) \emph{which} objective(s) an instruction refers to (condition type) from (ii) \emph{how strongly} each objective should be satisfied (condition intensity). 
As a result, compositional instructions can be mapped to entangled representations, making it difficult for the downstream policy to reliably interpret compound constraints.

\subsection{Multi-objective Representation Learning}

Prompt-driven control fundamentally relies on learning an embedding space that captures prompt semantics in a form usable for downstream generation or control. Sentence-level representations from pretrained language models~\cite{bert,reimers2019sentence} and multimodal aligned embeddings such as CLIP~\cite{radford2021learning} have shown that natural language can be mapped into continuous vectors that function as effective conditioning signals. However, when a prompt contains multiple attributes or constraints, a single monolithic embedding can suffer from semantic entanglement, where distinct semantic factors interfere and degrade the interpretability and reliability of compositional prompts~\cite{li2025geometric}.

To address this, recent representation learning efforts aim to decompose prompt semantics into structured factors and selectively emphasize relevant components, e.g., via disentangled or compositional representations in multimodal prompting and composition learning~\cite{hao2023learning,rahman2025dimple}. Related evidence from compositional multi-label semantics also suggests that explicitly modeling which semantic components are active improves generalization under combined meanings~\cite{chai2024compositional}. Following this perspective, we treat multi-condition instructions as a prompt-understanding problem and aim to structure the latent space so that \emph{condition type} (what) and \emph{condition intensity} (how much) are represented without undesired entanglement.

\section{Language-based Level Generation Objective}

\subsection{2D Level Generation Objective}
We adopt a widely used two-dimensional level-generation Gym environment for PCGRL research~\cite{pcgrl,cpcgrl}.
Our experiments use a modified \textit{Dungeon} setting with three tile types: \textit{Empty} \includegraphics[height=0.8em]{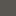}, \textit{Wall} \includegraphics[height=0.8em]{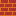}, and \textit{Bat} \includegraphics[height=0.8em]{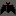}, instantiated with the \textit{Turtle} representation.
At each step, the observation encodes the current tile configuration (tile-type locations) together with the agent’s modification cursor.
The discrete action space consists of seven actions: four move the cursor to a neighboring location, and the remaining three assign the tile type (\textit{Empty}, \textit{Wall}, or \textit{Bat}) at the cursor position.
We fix the level resolution to \(16 \times 16\) and set the episode horizon to 1{,}500 steps.
Each episode starts from a randomly initialized grid sampled from the three-tile types.
During an episode, the agent repeatedly navigates and edits tiles to produce a final level that satisfies the instruction-specified constraints; progress toward the intended properties is reflected in the objective-defined reward signals.

We consider five controllable objectives $\tau_i$: region, path length, wall count, bat count, and bat direction.
\textit{\textbf{Region (RG)}} specifies the desired number of connected components in the traversable space, separated by walls.
\textit{\textbf{Path Length (PL)}} specifies a target distance for the longest-path between reachable locations within a connected region.
\textit{\textbf{Wall Count (WC)}} controls the total number of wall tiles.
\textit{\textbf{Bat Count (BC)}} controls the total number of bat tiles.
\textit{\textbf{Bat Direction (BD)}} controls the spatial distribution of bat tiles with respect to the four cardinal directions, encouraging bats to appear predominantly in a specified direction.
The first four objectives---\textit{RG}, \textit{PL}, \textit{WC}, and \textit{BC}---are conditioned on numerical target values $c$, whereas \textit{BD} is conditioned on one of four discrete direction labels.
Both the number of regions and the path length are computed using a flood-fill-based path-finding algorithm on the grid, with wall tiles treated as obstacles.

In this environment, rewards are defined in a \emph{delta} form: at each step, the agent receives feedback proportional to how much the current edit improves the objective relative to the previous state. Accordingly, each objective $\tau$ is associated with its own reward function $R_{\tau}$, and the step reward is computed from the change in the corresponding objective metric between consecutive states.

\subsection{Multi-Objective Instruction}
\label{sec:multi-objective_instruction}
\newcommand{\hlbox}[2]{%
  {\setlength{\fboxsep}{0pt}\colorbox{#1}{#2}}%
}

\newcommand{\hlRG}[1]{\hlbox{violet!20}{#1}}
\newcommand{\hlPL}[1]{\hlbox{blue!18}{#1}}
\newcommand{\hlWC}[1]{\hlbox{orange!20}{#1}}
\newcommand{\hlBC}[1]{\hlbox{green!20}{#1}}
\newcommand{\hlBD}[1]{\hlbox{red!18}{#1}}

\newcommand{\hlnumRG}[1]{\hlbox{violet!20}{#1}}
\newcommand{\hlnumPL}[1]{\hlbox{blue!18}{#1}}
\newcommand{\hlnumWC}[1]{\hlbox{orange!20}{#1}}
\newcommand{\hlnumBC}[1]{\hlbox{green!20}{#1}}
\newcommand{\hlnumBD}[1]{\hlbox{red!18}{#1}}

Representative examples for all instruction types are provided in Table~\ref{tab:instruction}.
The rightmost columns report the instruction-derived condition scalars $c_{i}$, each specifying the target value for objective $\tau_i$.
Entries are populated only for objectives activated by the instruction; inactive objectives are left blank.
The table also includes a color-coded legend that assigns a unique color to each objective (\hlRG{RG}, \hlPL{PL}, \hlWC{WC}, \hlBC{BC}, \hlBD{BD}).
The same colors highlight objective-relevant spans in the \textit{Instruction} column and mark the corresponding $c_{i}$ column.
This visualization explicitly links each linguistic cue to its objective and target value.
The source code and the instruction sets are available in this repository \footnote{\url{https://github.com/k-shyun/MIPCGRL}}.

\begin{table}[!t]
    \centering
    \caption{The example of language instruction set.}
    \vspace{-5pt}
    \begin{tabular}{m{0.6cm}m{0.9cm}m{3.2cm}m{0.15cm}m{0.15cm}m{0.15cm}m{0.15cm}m{0.15cm}}
    \toprule
    Type & Obj. & Instruction & $c_{\scriptscriptstyle\text{RG}}$ & $c_{\scriptscriptstyle\text{PL}}$ & $c_{\scriptscriptstyle\text{WC}}$ & $c_{\scriptscriptstyle\text{BC}}$ & $c_{\scriptscriptstyle\text{BD}}$ \\
    \midrule
    \multirow{8}{*}{Single}& $\tau_{\scriptscriptstyle\text{RG}}$ & \hlRG{\textit{A few}}-regions are present. & \hlnumRG{25} &  &  &  &  \\
    & $\tau_{\scriptscriptstyle \text{PL}}$ & The path is \hlPL{\textit{extended}} and \hlPL{\textit{prolonged}}, requiring considerable traversal effort. &  & \hlnumPL{60} &  &  &  \\
    & $\tau_{\scriptscriptstyle \text{WC}}$ & \hlWC{\textit{Overcrowded}} block area. &  &  & \hlnumWC{160} &  &  \\
    & $\tau_{\scriptscriptstyle \text{BC}}$ & \hlBC{\textit{A Few}} bats are scattered across the map. &  &  &  & \hlnumBC{10} &  \\
    & $\tau_{\scriptscriptstyle \text{BD}}$ & Bats spread at the \hlBD{\textit{bottom}}. &  &  &  &  & \hlnumBD{3} \\
    \midrule
    \multirow{17}{*}{Multi}& $\tau_{\scriptscriptstyle \text{WC}\oplus{}\text{BD}}$ &
      \hlWC{\textit{Overcrowded}} block area, \hlBD{\textit{Top-focused}} bat distribution.
      &  &  & \hlnumWC{160} &  & \hlnumBD{1} \\
    & $\tau_{\scriptscriptstyle \text{BC}\oplus{}\text{BD}}$ &
      \hlBC{\textit{Few}} bats, Bats grouped at the \hlBD{\textit{top}}.
      &  &  &  & \hlnumBC{10} & \hlnumBD{1} \\
    & $\tau_{\scriptscriptstyle \text{RG}\oplus{}\text{BC}}$ &
      The map has \hlRG{\textit{some}} regions, \hlBC{\textit{Several}} bats.
      &  \hlnumRG{50}&  &  & \hlnumBC{70} &  \\
    & $\tau_{\scriptscriptstyle \text{PL}\oplus{}\text{BC}}$ &
      \hlPL{\textit{Extended}} path length, a \hlBC{\textit{substantial number}} of monsters are scattered across the map.
      &  & \hlnumPL{60} &  & \hlnumBC{50} &  \\
    & $\tau_{\scriptscriptstyle \text{PL}\oplus{}\text{BD}}$ &
      \hlPL{\textit{Significant}} path length, The bats are gathered in the \hlBD{\textit{north}}, heavily occupying the \hlBD{\textit{top}} section.
      &  & \hlnumPL{60} &  &  & \hlnumBD{1} \\
    & $\tau_{\scriptscriptstyle \text{PL}\oplus{}\text{BC}\oplus{}\text{BD}}$ &
      \hlPL{\textit{Micro}} path length, \hlBC{\textit{Small}} bat group, Bats in the \hlBD{\textit{south}}.
      &  & \hlnumPL{20} &  & \hlnumBC{10} & \hlnumBD{3} \\
    \bottomrule
    \end{tabular}
    \label{tab:instruction}
    \vspace{-15pt}
\end{table}

\textbf{Single-objective Instruction Set.}
We construct a single-objective instruction dataset consisting of 80 sentences (29.7$\pm{}$21.10 characters), where each sentence specifies \emph{one} generation objective $\tau$ (among the five objectives) and its target \emph{condition value} $c$ used for conditioning.
We denote the instruction set for objective $\tau_{i}$ (e.g., $\tau_{\text{RG}}$).
In addition, to evaluate compositional generalization without linguistic rewriting, we form a concatenated test setting by appending two single-objective instructions, denoted as $\tau_{i_1+i_2}$ (e.g. $\tau_{\text{RG}+\text{BC}}$); the model is trained on $\tau_{i_1}$ and $\tau_{i_2}$ separately, and concatenation is applied only at test time.

\textbf{Multi-Objective Instruction Set.}
To evaluate compositional control, we additionally build a multi-objective instruction dataset of 1280 sentences (61.1$\pm{}$32.08 characters).
Each multi-objective instruction contains \emph{two} objectives selected from the five objectives and specifies a target condition value for each objective (e.g., $\tau_{\text{RG}\oplus\text{BC}}$), requiring the model to interpret and satisfy multiple constraints described within a single sentence.
While the main multi-objective instruction set consists of two-objective compositions, Section~\ref{sec:exp_add_task} further considers triple-objective instructions (e.g., $\tau_{\text{PL}\oplus\text{BC}\oplus\text{BD}}$) constructed in the same manner to examine generalization to subset-objective instructions.

\begin{figure*}[!t]
    \centering

    \includegraphics[width=1\linewidth]{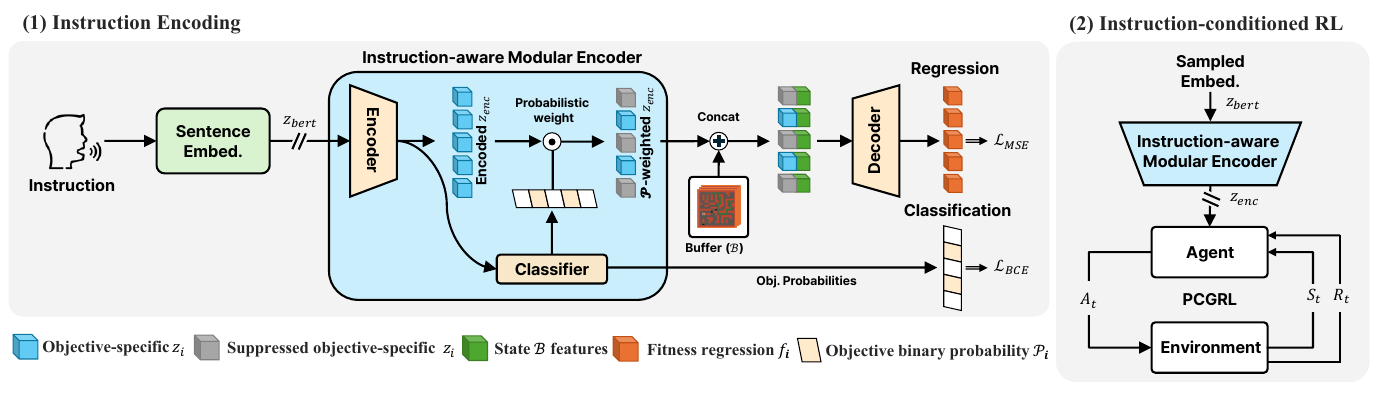}
    \vspace{-0.6cm}
    \caption{\textbf{Overview of the proposed MIPCGRL framework}, which comprises two main stages: (1) training an objective-specific instruction encoder that separates instructions into objective-specific representations using a classifier, multi-head regression, and a probabilistic weighting mechanism, and (2) training an RL agent conditioned on the encoded instructions. The encoder is pre-trained with both regression and classification objectives and reused during RL training. Dashed lines indicate a gradient stop.
    }
    \label{fig:architecture}
    \vspace{-0.5cm}
\end{figure*}

\section{Methodology}

We propose MIPCGRL, an instruction-aware multi-objective procedural content generation framework that enhances the generalization and controllability of IPCGRL. The key idea is to encode natural language instructions into objective-specific representations that disentangle multiple generation objectives, thereby enabling multi-objective decision-making in RL. As shown in Fig. \ref{fig:architecture}, MIPCGRL adopts a two-stage training framework. In the first stage, an instruction–objective encoder maps each instruction to objective-level embeddings using a language encoder and is trained to distinguish different generation objectives. In the second stage, the learned encoder is fixed and used to condition a DRL agent during training, as detailed in the subsequent sections.

\subsection{Instruction-aware Modular Representation}

A natural-language instruction $\mathcal{I}$ is embedded using a pre-trained $\text{BERT}_\psi$\footnote{\url{https://huggingface.co/google-bert/bert-base-uncased}}.
The final-layer [CLS] token is used as a sentence-level embedding, denoted by $z_{\mathrm{bert}}=\text{BERT}_\psi(\mathcal{I})$.
The backbone $\text{BERT}_\psi$ is kept frozen, and an encoder $E_\theta$ maps $z_{\mathrm{bert}}$ to a compressed latent vector
$z_{\mathrm{enc}}=E_\theta(z_{\mathrm{bert}})\in\mathbb{R}^{n_{\mathrm{obj}}\cdot d}$,
where $d$ is the dimensionality of each objective-specific representation and $n_{\mathrm{obj}}$ is the number of predefined objectives.
The latent vector $z_{\mathrm{enc}}$ is then processed by two parallel modules: (1) a multi-label objective classifier and (2) a multi-head fitness regressor.

\subsubsection{Multi-label Objective Classifier (CLS)}
The classifier $C_\theta$ predicts which objectives are semantically active in $\mathcal{I}$:
\begin{equation}
[\mathcal{P}_{\text{RG}},\mathcal{P}_{\text{PL}},\mathcal{P}_{\text{WC}},\mathcal{P}_{\text{BC}},\mathcal{P}_{\text{BD}}]
= C_\theta(z_{\mathrm{enc}}),
\label{eq:cls}
\end{equation}
where $\mathcal{P}_\tau$ denotes the probability of activating objective $\tau$.
During training, each objective is annotated with a binary label $y_\tau\in\{0,1\}$ indicating presence in the instruction, and $C_\theta$ is optimized with the binary cross-entropy loss $\mathcal{L}_{\mathrm{BCE}}$.
These probabilities are used to gate objective-wise latent factors via probabilistic weighting.

\subsubsection{Multi-head Fitness Regression (REG)}
The latent vector $z_{\mathrm{enc}}\in\mathbb{R}^{n_{\mathrm{obj}}\cdot d}$ is reshaped into an objective-wise latent matrix
$z^{\mathrm{obj}}\in\mathbb{R}^{n_{\mathrm{obj}}\times d}$, whose $i$-th row $z^{\mathrm{obj}}_{\tau}\in\mathbb{R}^{d}$ encodes semantics relevant to $i$-th objective 
(e.g., $[z_{\text{RG}}, z_{\text{PL}}, z_{\text{WC}}, z_{\text{BC}}, z_{\text{BD}}]$).
Each objective representation is scaled by the corresponding classifier probability from Eq.~\ref{eq:cls}:
\begin{equation}
z^{weighted}_{\tau}=\mathcal{P}_\tau\, z^{\mathrm{obj}}_{\tau}.
\label{eq:weighted_z}
\end{equation}
This probabilistic weighting retains representations associated with semantically active objectives while suppressing irrelevant ones, yielding the final weighted objective representation $z^{weighted}$. The same operation is applied during both training and inference.

The weighted representation $z^{weighted}$ obtained via Eq.~\ref{eq:weighted_z} is concatenated with a sampled state $s'\sim\mathcal{B}$ from the offline state dataset $\mathcal{B}$,
and passed to the decoder $D_\theta$ to predict objective-wise fitness values:
\begin{equation}
[\hat f_{\text{RG}}, \hat f_{\text{PL}}, \hat f_{\text{WC}}, \hat f_{\text{BC}}, \hat f_{\text{BD}}]
= D_\theta([z^{weighted}, s']).
\label{eq:fitness_pred}
\end{equation}

\noindent\textbf{Loss functions.}
Let $\mathcal{T}=\{\tau_{\mathrm{RG}},\tau_{\mathrm{PL}},\tau_{\mathrm{WC}},\tau_{\mathrm{BC}},\tau_{\mathrm{BD}}\}$ denote the predefined objective set.
For each $\tau\in\mathcal{T}$, let $y_\tau\in\{0,1\}$ indicate whether objective $\tau$ is active in the instruction, and let $P_\tau$ be the corresponding classifier output in Eq.~\ref{eq:cls}.
The predictions $\hat f_\tau$ in Eq.~\ref{eq:fitness_pred} are supervised using target fitness scores $f_\tau$, where each score is computed by an objective-specific fitness function $F_{\tau}(s',c_\tau)$ that quantifies how well the sampled state $s'$ satisfies the target condition $c_\tau$ specified by the instruction \cite{ipcgrl}.
The multi-label classification loss and the multi-head regression loss are defined as
\begin{align}
\mathcal{L}_{\mathrm{BCE}}
&= -\sum_{\tau\in\mathcal{T}}\!\left[y_\tau \log P_\tau + (1-y_\tau)\log(1-P_\tau)\right], \label{eq:lbce}\\
\mathcal{L}_{\mathrm{MSE}}
&= \frac{1}{\sum_{\tau\in\mathcal{T}} y_{\tau}}
\sum_{\tau\in\mathcal{T}} y_{\tau}\,\|\hat f_{\tau}- f_{\tau}\|_2^2, \label{eq:lmse}
\end{align}
As shown in Eq.~\ref{eq:lmse}, the regression loss is computed only over active objectives ($y_\tau=1$) and normalized by the number of active objectives.
The overall objective is defined as the sum of the classification and regression losses in Eqs.~\ref{eq:lbce} and~\ref{eq:lmse}:
\begin{equation}
\mathcal{L}_{\mathrm{total}}=\mathcal{L}_{\mathrm{BCE}}+\mathcal{L}_{\mathrm{MSE}}.
\label{eq:ltotal}
\end{equation}

\newcommand{\basecell}[1]{\cellcolor{gray!15}{#1}}

\begin{table*}[t]
\caption{\textbf{Comprehensive results across objective combinations.}\\ Relative performance is measured against a text-based generator baseline using the average progress over the two objectives.}
\vspace{-5pt}
\centering
\footnotesize
\begin{tabular}{C{1.5cm}L{1.6cm}L{1.2cm}C{1.6cm}C{1.6cm}C{1.8cm}C{1.6cm}C{1.6cm}C{1.8cm}}

\toprule
 &  & & \multicolumn{3}{c}{Single-objective ($\tau_{a+b}$)} &
       \multicolumn{3}{c}{Multi-objective ($\tau_{a\oplus b}$)} \\

Obj. ($\tau$)& \multicolumn{1}{c}{Method} & \multicolumn{1}{c}{Input Type} &
$P_{\tau_a}$ & $P_{\tau_b}$ & Improvement &
$P_{\tau_a}$ & $P_{\tau_b}$ & Improvement \\

\midrule

\multirow{4}{*}{$\tau_{\text{WC}}$, $\tau_{\text{BD}}$} & \graycell Cont &  \graycell Scalar
& \graycell 68.73{\scriptsize$\pm$4.03} & \graycell 66.02{\scriptsize$\pm$1.25} & \graycell -
& \graycell 57.11{\scriptsize$\pm$1.12} & \graycell 76.58{\scriptsize$\pm$0.98} & \graycell - \\
& \graycell Cont (goal) & \graycell Scalar
& \graycell 69.70{\scriptsize$\pm$4.89} & \graycell 65.56{\scriptsize$\pm$1.88} & \graycell -
& \graycell 55.17{\scriptsize$\pm$1.48} & \graycell \textbf{77.08{\scriptsize$\pm$2.25}} & \graycell - \\
\cline{2-9}\addlinespace[0.05cm]
& IPCGRL & Text
& 69.05{\scriptsize$\pm$1.89} & 64.57{\scriptsize$\pm$2.13} & -
& 73.68{\scriptsize$\pm$1.42} & 59.11{\scriptsize$\pm$1.74} & - \\
& \skycell \textbf{MIPCGRL} & \skycell Text
& \skycell \textbf{77.03{\scriptsize$\pm$6.35}} & \skycell \textbf{68.32{\scriptsize$\pm$1.02}} & \skycell \textcolor{green!60!black}{+8.8\%}
& \skycell \textbf{77.34{\scriptsize$\pm$3.47}} & \skycell 72.18{\scriptsize$\pm$2.37} & \skycell \textcolor{green!60!black}{+12.6\%}* \\

\midrule

\multirow{4}{*}{$\tau_{\text{BC}}$, $\tau_{\text{BD}}$} & \graycell Cont &  \graycell Scalar
& \graycell 43.78{\scriptsize$\pm$3.37} & \graycell 75.17{\scriptsize$\pm$3.55} & \graycell -
& \graycell 66.66{\scriptsize$\pm$1.21} & \graycell 51.78{\scriptsize$\pm$0.64} & \graycell - \\
& \graycell Cont (goal) & \graycell Scalar
& \graycell 57.50{\scriptsize$\pm$11.47} & \graycell 72.27{\scriptsize$\pm$9.04} & \graycell -
& \graycell 63.34{\scriptsize$\pm$6.06} & \graycell 62.92{\scriptsize$\pm$12.05} &\graycell  - \\
\cline{2-9}\addlinespace[0.05cm]
& IPCGRL& Text
& 51.91{\scriptsize$\pm$20.70} & 76.12{\scriptsize$\pm$3.83} & -
& \textbf{76.89{\scriptsize$\pm$2.73}} & 71.84{\scriptsize$\pm$5.97} & - \\
& \skycell \textbf{MIPCGRL} & \skycell Text
& \skycell \textbf{78.33{\scriptsize$\pm$4.48}} & \skycell \textbf{76.81{\scriptsize$\pm$0.40}} & \skycell \textcolor{green!60!black}{+21.2\%}
& \skycell 66.61{\scriptsize$\pm$1.47} & \skycell \textbf{86.48{\scriptsize$\pm$2.39}} & \skycell \textcolor{green!60!black}{+2.9\%} \\

\midrule

\multirow{4}{*}{$\tau_{\text{RG}}$, $\tau_{\text{BC}}$} & \graycell Cont & \graycell Scalar
& \graycell 31.79{\scriptsize$\pm$15.42} & \graycell 50.96{\scriptsize$\pm$3.66} & \graycell -
& \graycell 34.37{\scriptsize$\pm$ 1.61} & \graycell 61.73{\scriptsize$\pm$1.10} & \graycell - \\
& \graycell Cont (goal) & \graycell Scalar
& \graycell 26.05{\scriptsize$\pm$13.52} & \graycell 52.46{\scriptsize$\pm$1.80} & \graycell -
& \graycell 33.02{\scriptsize$\pm$0.54} & \graycell 63.92{\scriptsize$\pm$0.36} & \graycell - \\
\cline{2-9}\addlinespace[0.05cm]
& IPCGRL & Text
& 61.52{\scriptsize$\pm$5.13} & \textbf{73.98{\scriptsize$\pm$2.45}} & -
& \textbf{41.63{\scriptsize$\pm$3.05}} & 51.89{\scriptsize$\pm$0.76} & - \\
& \skycell \textbf{MIPCGRL} & \skycell Text
& \skycell \textbf{62.20{\scriptsize$\pm$7.81}} & \skycell 71.48{\scriptsize$\pm$1.46} & \skycell \textcolor{red!60!black}{-1.3\%}
& \skycell 41.56{\scriptsize$\pm$5.43} & \skycell \textbf{69.64{\scriptsize$\pm$1.97}} & \skycell \textcolor{green!60!black}{+18.9\%}* \\

\midrule

\multirow{4}{*}{$\tau_{\text{PL}}$, $\tau_{\text{BC}}$} & \graycell Cont & \graycell Scalar
& \graycell 30.76{\scriptsize$\pm$2.08} & \graycell 52.50{\scriptsize$\pm$1.89} & \graycell -
& \graycell 39.92{\scriptsize$\pm$1.86} & \graycell 69.78{\scriptsize$\pm$2.02} & \graycell - \\
& \graycell Cont (goal) & \graycell Scalar
& \graycell 35.38{\scriptsize$\pm$6.28} & \graycell 60.66{\scriptsize$\pm$7.82} & \graycell -
& \graycell 38.12{\scriptsize$\pm$1.60} & \graycell 72.07{\scriptsize$\pm$1.38} & \graycell - \\
\cline{2-9}\addlinespace[0.05cm]
& IPCGRL & Text
& 33.26{\scriptsize$\pm$3.76} & 77.40{\scriptsize$\pm$7.60} & -
& \textbf{44.07{\scriptsize$\pm$1.17}} & 45.52{\scriptsize$\pm$6.34} & - \\
& \skycell \textbf{MIPCGRL} & \skycell Text
& \skycell \textbf{39.16{\scriptsize$\pm$6.11}} & \skycell \textbf{80.55{\scriptsize$\pm$5.30}} & \skycell \textcolor{green!60!black}{+8.2\%}
& \skycell 42.31{\scriptsize$\pm$1.61} & \skycell \textbf{75.25{\scriptsize$\pm$1.84}} & \skycell \textcolor{green!60!black}{+31.2\%}* \\

\midrule

\multirow{4}{*}{$\tau_{\text{PL}}$, $\tau_{\text{BD}}$} & \graycell Cont & \graycell Scalar
& \graycell 35.55{\scriptsize$\pm$2.72} & \graycell 64.80{\scriptsize$\pm$0.42} & \graycell -
& \graycell 25.66{\scriptsize$\pm$4.02} & \graycell 77.68{\scriptsize$\pm$5.13} & \graycell - \\
& \graycell Cont (goal) & \graycell Scalar
& \graycell 28.95{\scriptsize$\pm$4.82} & \graycell 65.12{\scriptsize$\pm$1.18} & \graycell -
& \graycell 24.58{\scriptsize$\pm$2.46} & \graycell 71.97{\scriptsize$\pm$2.53} & \graycell - \\
\cline{2-9}\addlinespace[0.05cm]
& IPCGRL & Text
& 40.16{\scriptsize$\pm$3.16} & 67.87{\scriptsize$\pm$1.42} & -
& \textbf{32.29{\scriptsize$\pm$7.40}} & 65.90{\scriptsize$\pm$1.66} & - \\
& \skycell \textbf{MIPCGRL} & \skycell Text
& \skycell \textbf{45.87{\scriptsize$\pm$3.87}} & \skycell \textbf{69.44{\scriptsize$\pm$1.62}} & \skycell \textcolor{green!60!black}{+6.7\%}
& \skycell 25.20{\scriptsize$\pm$0.4}5 & \skycell \textbf{80.57{\scriptsize$\pm$0.78}} & \skycell \textcolor{green!60!black}{+7.7\%} \\

\bottomrule
\end{tabular}
\par\smallskip
{\footnotesize \makebox[\linewidth][r]{* indicates $p<0.05$ (Welch's $t$-test) over\ IPCGRL.}}
\vspace{-22pt}
\label{tab:single_multi_result}
\end{table*}

\subsection{Instruction-conditioned RL}

During RL training and deployment, the trained encoder takes a natural language instruction \(\mathcal{I}\) as input and generates \(z^{weighted}\), which conditions the RL policy; the encoder is kept fully frozen, and gradients from the RL loss are stopped at \(z^{weighted}\). At each timestep \(t\), the agent receives a state \(s_t\), samples an action \(a_t=\pi(s_t, z^{weighted})\), and transitions to the next state \(s_{t+1}=f(s_t,a_t)\). The instruction-conditioned latent vector is held fixed throughout the rollout, guiding the policy according to the specified multi-objective instruction.

To provide an instruction-aligned learning signal under this fixed conditioning, let $\mathcal{T}$ denote the predefined objective set.
Given an instruction $\mathcal{I}$, a mapping function $\Phi$ maps it to the set of active objective–-condition pairs, where each $(\tau,c_\tau)\in\Phi(\mathcal{I})$ indicates that objective $\tau$ is active and is associated with the target condition value $c_\tau$ specified by $\mathcal{I}$.
For instance, an instruction $\mathcal{I}$ such as  "\textit{\hlPL{Significant path length}, \hlBC{A Few bats are scattered across the map.}}" is mapped to {\setlength{\fboxsep}{0.5pt}%
$\Phi(\mathcal{I})=[(\colorbox{blue!15}{$\tau_{\text{PL}}, 60$}),\, (\colorbox{green!15}{$\tau_{\text{BC}}, 10$})]$.}
For each objective $\tau\in\mathcal{T}$, an objective-specific reward component $R_\tau(s_t,s_{t+1},c_\tau)$ is defined.
The unified scalar reward used for RL optimization is then
\begin{equation}
R(s_t,s_{t+1},\mathcal{I})=\sum_{(\tau,c_\tau)\in\Phi(\mathcal{I})} w_\tau\,R_\tau(s_t,s_{t+1},c_\tau),
\label{eq:weighted_reward}
\end{equation}
where $w_\tau$ is an objective-specific weighting factor controlling the relative contribution among the active objectives.
By summing only over $(\tau,c_\tau)\in\Phi(\mathcal{I})$, Eq.~\ref{eq:weighted_reward} ensures that the policy is optimized exclusively with respect to the objectives specified by the instruction, while conditioning each component on $c_\tau$ maintains trajectory-level consistency with the instruction-defined targets throughout the rollout.

\section{Experiment Setup}
\subsection{Model Training}
\label{sec: model training}
The instruction encoder is trained for 100 epochs using objective-wise normalized target fitness values $f_\tau$, where the normalized targets are linearly scaled to the range [$-5, 5$].
For the encoder training, we use a state buffer $\mathcal{B}$ containing approximately 86K unique states collected from a controllable PCGRL agent \cite{cpcgrl}.
To promote diverse state coverage, duplicate states are removed during buffer construction.
The RL policy is trained using Proximal Policy Optimization (PPO)\cite{schulman2017proximal} for a total of 50 million timesteps implemented in PureJaxRL \cite{purejaxrl}, and executed on RTX 8000 GPU machines.
The RL hyperparameters are set as follows: $10$ epoch size, $128$ rollout length, $\gamma_{\text{GAE}} = 0.95$, $\gamma = 0.99$, $10^{-4}$ learning rate.
The results were averaged over three independent runs with different random seeds.
For each template-based instruction in Section~\ref{sec:multi-objective_instruction}, we sampled 10 level instances, resulting in 160 samples per objective per seed for the single-objective setting and 2,540 samples per objective pair per seed for the multi-objective setting.
To address reward scale imbalance, we assign weights as follows: $w_{\text{PL}}=1;w_{\text{RG}}, w_{\text{WC}},w_{\text{BC}}, w_{\text{BD}}=0.15$.
For all experiments, we use the JAX-based GPU-accelerated implementation of the PCGRL framework introduced in \cite{pcgrl+}.

\subsection{Evaluation Metric} The \textit{Progress} \cite{ipcgrl} evaluates how the level satisfies the condition in the instruction.
The metric values the performance as $1 - | \frac{g-s_T}{g-s_0}|$, where $g$ is the goal condition, $s_T$ and $s_0$ are the terminal- and initial state status, respectively. The score represents normalized progress toward the target condition and is scaled to the range [0, 100].

\subsection{Comparison}
\textbf{Text-based baseline.}
IPCGRL~\cite{ipcgrl} is adopted as the primary text-conditioned baseline for evaluating multi-objective generation in language-instructed PCGRL. 
Prior results show that IPCGRL performs strongly under single-objective instructions, but degrades under multi-objective compositions, making it a suitable reference for assessing multi-objective generalization and robustness.

\textbf{Scalar-based baselines.}
We also include a controllable PCGRL baseline, denoted as Cont~\cite{cpcgrl}, as a non-text, scalar-conditioned reference.
Cont uses a directional control signal \( c = \mathrm{sign}(g - f(s_t)) \) to guide generation, while its goal-conditioned variant, Cont(goal)~\cite{ipcgrl}, directly uses the target goal as the conditioning input, i.e., \( c_t = g \).
These scalar-conditioned baselines are included to ensure a fair comparison and to provide performance reference points.

\section{Experimental Results}
\label{sec:experiment}

This section investigates whether MIPCGRL mitigates the above limitation by retaining strong performance in single-objective settings while improving robustness under multi-objective instruction settings.
We evaluate MIPCGRL in two aspects:  
(1) representation capability for single- and multi-objective instructions, assessing whether single-objective performance is preserved and multi-objective performance is improved (Section~\ref{sec:exp_single_multi});  and
(2) representation capability for subset-objective instructions, assessing whether representations learned under multi-objective compositions generalize to previously unseen subset-objective instructions (Section~\ref{sec:exp_add_task}).

\newcommand{\twoimg}[2]{%
  \raisebox{-.05\height}{%
    \raisebox{8.5ex}{1}%
    \hspace{0.2em}%
    \includegraphics[width=1.40cm]{#1}%
    \hspace{0.2em}%
    \raisebox{8.5ex}{2}%
    \hspace{0.2em}%
    \includegraphics[width=1.40cm]{#2}%
  }%
}

\newcommand{\twoimgtf}[2]{%
  \raisebox{-.05\height}{%
    \raisebox{8.5ex}{3}%
    \hspace{0.2em}%
    \includegraphics[width=1.40cm]{#1}%
    \hspace{0.2em}%
    \raisebox{8.5ex}{4}%
    \hspace{0.2em}%
    \includegraphics[width=1.40cm]{#2}%
  }%
}
\begin{table}[!t]
\caption{\textbf{Qualitative examples of multi-objective generation.} Highlighted spans are annotated with $(\tau, c)$; representative text-conditioned generations are shown.
}
\vspace{-5pt}
\centering
{
\begin{tabular}{@{\hspace{2pt}} m{0.6cm} m{4.0cm} m{3.1cm} }
\toprule
\textbf{Obj.} & \textbf{Text} & \textbf{Level} \\
\multirow{5}{*}{$\tau_{\scriptscriptstyle \text{RG}\oplus{}\text{BC}}$}
&
1. \hlRG{\textit{A few{$\scriptscriptstyle \text{(RG, 25)}$}}}  regions,  \hlBC{\textit{Small{$\scriptscriptstyle \text{(BC, 10)}$}}} bat group.  \newline
2. \hlRG{\textit{A few{$\scriptscriptstyle \text{(RG, 25)}$}}}  regions, \hlBC{\textit{Several{$\scriptscriptstyle \text{(BC, 70)}$}}} bats.
&
\twoimg{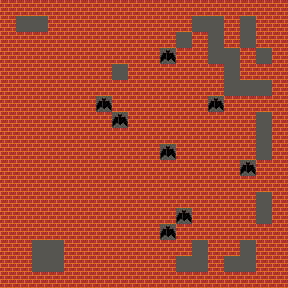}
       {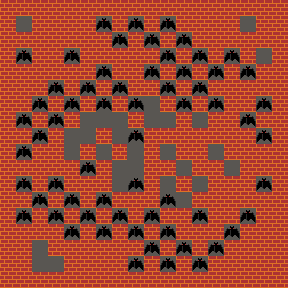}
\\

&
3. \hlRG{\textit{Moderately scattered{$\scriptscriptstyle \text{(RG, 75)}$}}}  regions,  \hlBC{\textit{Small{$\scriptscriptstyle \text{(BC, 10)}$}}} bat group.  \newline
4. \hlRG{\textit{Moderately scattered{$\scriptscriptstyle \text{(RG, 75)}$}}}  regions, \hlBC{\textit{Several{$\scriptscriptstyle \text{(BC, 70)}$}}} bats.
&
\twoimgtf{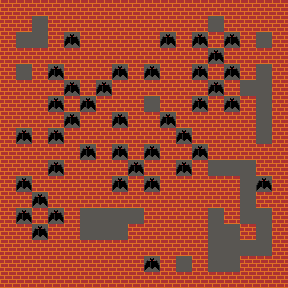}
       {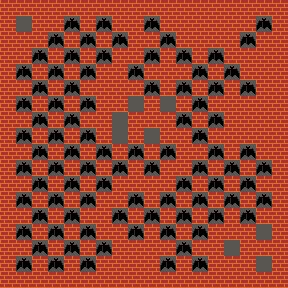} 
\\

\midrule
\multirow{5}{*}{$\tau_{\scriptscriptstyle \text{PL}\oplus{}\text{BC}}$}
&
1. \hlPL{\textit{Minimal{$\scriptscriptstyle \text{(PL, 20)}$}}} path length, \hlBC{\textit{Some{$\scriptscriptstyle \text{(BC, 40)}$}}} bats. \newline
2. \hlPL{\textit{Minimal{$\scriptscriptstyle \text{(PL, 20)}$}}} path length, Bat \hlBC{\textit{swarm{$\scriptscriptstyle \text{(BC, 100)}$}}}.
&
\twoimg{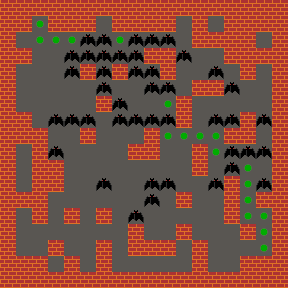}
       {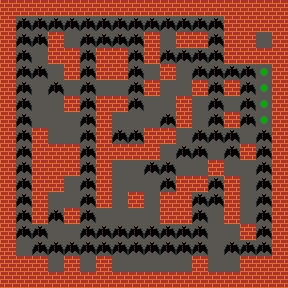}
\\

&
3. The path is of \hlPL{\textit{moderate{$\scriptscriptstyle \text{(PL, 60)}$}}} length, \hlBC{\textit{Some{$\scriptscriptstyle \text{(BC, 40)}$}}} bats. \newline
4. The path is of \hlPL{\textit{moderate{$\scriptscriptstyle \text{(PL, 60)}$}}} length,  Bat \hlBC{\textit{swarm{$\scriptscriptstyle \text{(BC, 100)}$}}}.
&
\twoimgtf{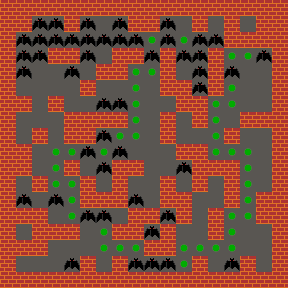}
       {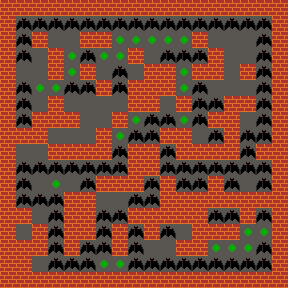}
\\

\midrule
\multirow{5}{*}{$\tau_{\scriptscriptstyle \text{WC}\oplus{}\text{BD}}$}
&
1. \hlWC{\textit{Few{$\scriptscriptstyle \text{(WC, 40)}$}}} blocks, \hlBD{\textit{Top-side{$\scriptscriptstyle \text{(BD, 1)}$}}} bat concentration. \newline
2. \hlWC{\textit{Few{$\scriptscriptstyle \text{(WC, 40)}$}}} blocks, \hlBD{\textit{Bottom-side{$\scriptscriptstyle \text{(BD, 3)}$}}} bat concentration.
&
\twoimg{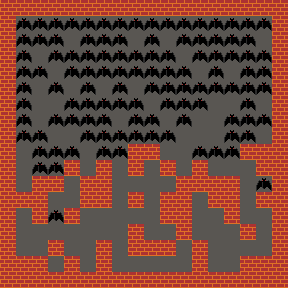}
       {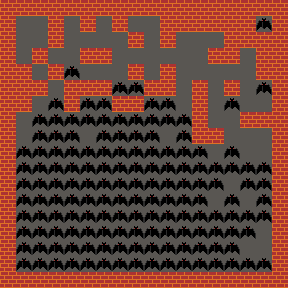}
\\

&
3. \hlWC{\textit{High-density{$\scriptscriptstyle \text{(WC, 160)}$}}} blocks, \hlBD{\textit{Top-side{$\scriptscriptstyle \text{(BD, 1)}$}}} bat concentration. \newline
4. \hlWC{\textit{High-density{$\scriptscriptstyle \text{(WC, 160)}$}}} blocks, \hlBD{\textit{Bottom-side{$\scriptscriptstyle \text{(BD, 3)}$}}} bat concentration.
&
\twoimgtf{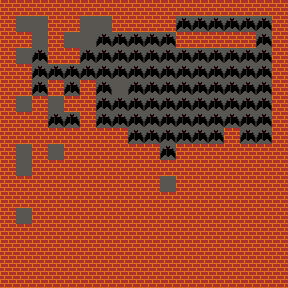}
       {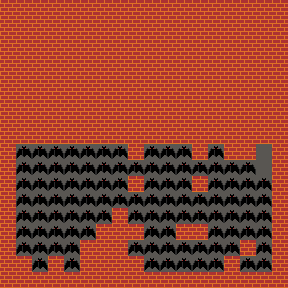}
\\

\bottomrule
\end{tabular}
\vspace{-15pt}
\label{tab:result_example}
}
\end{table}

\subsection{Multi-Objective Generation}
\label{sec:exp_single_multi}
IPCGRL exhibits clear limitations in mitigating objective interference and achieving reliable joint optimization under multi-objective instructions, whereas MIPCGRL consistently preserves objective-specific controllability and stabilizes optimization across diverse objective combinations. As shown in Table~\ref{tab:single_multi_result}, MIPCGRL yields substantial improvements in the multi-objective setting, achieving an average performance gain of 13.2\% over IPCGRL and 13.3\% over Cont (goal). These gains are particularly pronounced for challenging compositions, including $\tau_{\text{WC} \oplus \text{BD}}$, $\tau_{\text{RG} \oplus \text{BC}}$ and  $\tau_{\text{PL} \oplus \text{BC}}$, where the improvements over IPCGRL are statistically significant ($p < 0.05$). Similarly, MIPCGRL significantly outperforms Cont (goal) on $\tau_{\text{BC} \oplus \text{BD}}$ and $\tau_{\text{PL} \oplus \text{BC}}$  ($p < 0.05$), while maintaining consistent advantages across the remaining combinations.
Representative examples in Table~\ref{tab:result_example} further illustrate that MIPCGRL produces text-conditioned generations that more faithfully reflect the objective-specific conditions described in multi-objective instructions.

Importantly, the benefits observed under multi-objective settings do not come at the expense of single-objective performance. The same objective-aware representation enables MIPCGRL to preserve—and frequently improve—performance relative to IPCGRL in the single-objective setting (with an average improvement of 8.7\%), outperforming it in four out of five objectives with no statistically significant degradation in the remaining case. Moreover, MIPCGRL achieves an average improvement of 25.4\% over Cont (goal), with statistically significant gains on $\tau_{\text{RG}+\text{BC}}$ and $\tau_{\text{PL}+\text{BD}}$ ($p < 0.05$). Taken together, these results indicate that MIPCGRL enhances objective awareness and robustness for complex multi-objective instructions without introducing adverse trade-offs, providing a reliable framework for multi-objective instruction-conditioned optimization.

\begin{figure*}[!t]
    \centering
    \includegraphics[width=1\linewidth]{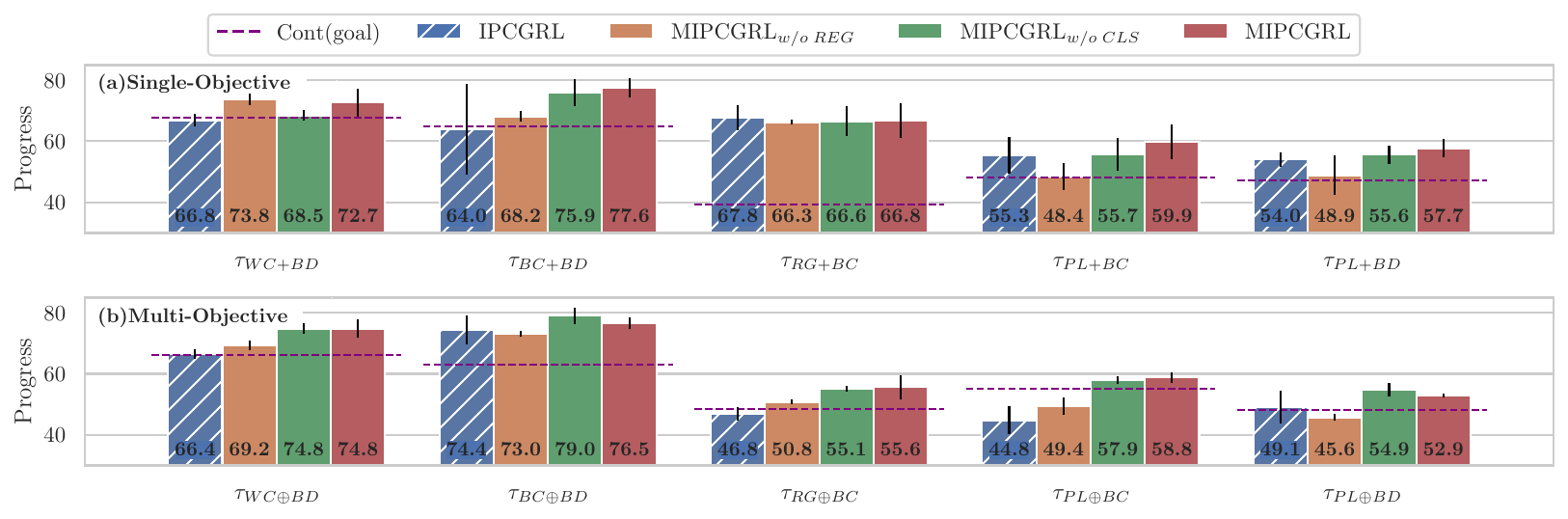}
    \vspace{-0.6cm}
    \caption{We conducted experiments across two instruction settings:
    (a) Single-Objective Composition
    (b) Multi-Objective Composition.
    In all configurations, we evaluated the baseline models Cont(goal) and IPCGRL, the proposed model MIPCGRL, and conducted ablation studies by removing the multi-head regression (REG) and objective classifier (CLS) components.
    }
    \label{fig:result graph}
    \vspace{-18pt}
\end{figure*}

\subsection{Compositional Generalization to Unseen Objective Subsets}
\label{sec:exp_add_task}
\begin{table}[!t]
\caption{\textbf{Generalization to objective-subset settings.}\\Gray rows indicate the in-distribution upper bound.}
\vspace{-5pt}
\centering
\begin{tabular}{p{1.1cm} p{1.1cm} c c c}
\toprule
 \multicolumn{2}{c}{Objective}& \multicolumn{3}{c}{Progress} \\
\multicolumn{1}{c}{\textbf{Train}} & \multicolumn{1}{c}{\textbf{Test}} &   $P_{\tau_\text{PL}}$ & $P_{\tau_\text{BC}}$ & $P_{\tau_\text{BD}}$  \\

\midrule
\multirow{4}{*}{$\tau _{\scriptscriptstyle \text{PL}\oplus \text{BC}\oplus \text{BD}}$} &
$\tau _{\scriptscriptstyle \text{PL}\oplus \text{BC}\oplus \text{BD}}$ & 
33.26{\scriptsize$\pm$1.27} & 56.13{\scriptsize$\pm$1.25} & 69.23{\scriptsize$\pm$1.46}  \\
\addlinespace[0.05cm]
 \cline{2-5}\addlinespace[0.05cm]
&
$\tau _{\scriptscriptstyle \text{PL}\oplus \text{BC}}$ & 
33.80{\scriptsize$\pm$1.41} & 67.52{\scriptsize$\pm$4.03} & - \\
 & 
$\tau _{\scriptscriptstyle \text{PL}\oplus \text{BD}}$ & 
34.47{\scriptsize$\pm$2.18} & - & 82.02{\scriptsize$\pm$1.72} \\
 & 
$\tau _{\scriptscriptstyle \text{BC}\oplus \text{BD}}$ & 
- & 42.63{\scriptsize$\pm$3.19} & 73.89{\scriptsize$\pm$1.55}  \\

\midrule

\graycell$\tau _{\scriptscriptstyle \text{PL}\oplus \text{BC}}$ & 
\graycell$\tau _{\scriptscriptstyle \text{PL}\oplus \text{BC}}$ &
\graycell42.31{\scriptsize$\pm$1.61} & \graycell75.25{\scriptsize$\pm$1.84}& \graycell- \\

\graycell$\tau _{\scriptscriptstyle \text{PL}\oplus \text{BD}}$ & 
\graycell$\tau _{\scriptscriptstyle \text{PL}\oplus \text{BD}}$ & 
\graycell25.20{\scriptsize$\pm$0.45} & \graycell- & \graycell80.57{\scriptsize$\pm$0.78} \\

\graycell$\tau _{\scriptscriptstyle \text{BC}\oplus \text{BD}}$ & 
\graycell$\tau _{\scriptscriptstyle \text{BC}\oplus \text{BD}}$ & 
\graycell- & \graycell66.61{\scriptsize$\pm$1.47} & \graycell86.48{\scriptsize$\pm$2.39} \\
\addlinespace[0.05cm]

\bottomrule

\end{tabular}
\vspace{-15pt}
\label{tab:add_task_result}
\end{table}

Training on every objective combination quickly becomes combinatorial, making it inefficient to cover all subset instructions explicitly; therefore, this experiment evaluates whether representations learned under more complex compositions transfer to unseen objective subsets. 
Specifically, a policy trained with the triple-objective instruction $\tau_{\text{PL} \oplus \text{BC} \oplus \text{BD}}$ is evaluated on two-objective subset instructions ($\tau_{\text{PL} \oplus \text{BC}}$, $\tau_{\text{PL} \oplus \text{BD}}$, and $\tau_{\text{BC} \oplus \text{BD}}$), which constitute unseen combinations of objectives relative to training.
This setting assesses whether the learned representation supports selective compositionality—i.e., suppressing the removed objective while preserving and jointly optimizing the remaining ones—thereby revealing robustness to distributional shifts in instruction structure and objective-wise disentanglement.

As shown in Table~\ref{tab:add_task_result}, a policy trained with triple-objective instructions generalizes robustly to instruction reduction. 
Two-objective subset instructions correspond to previously unseen combinations of objectives, yet the objective progress remains in a similar performance regime to in-distribution triple-objective evaluation.
Differences across subsets reflect a redistribution of progress across objectives, rather than a degradation of overall behavior. 
Compared to policies trained directly on the corresponding two-objective instructions (gray rows), the triple-objective policy shows some performance loss on subset evaluations.
Nevertheless, its optimization remains stable without severe across-objective deterioration.
Overall, these results suggest that multi-objective training preserves objective-relevant factors that can be selectively reused under subset instructions, supporting improved compositional generalization.

\section{Discussion}
\subsection{Quantitative Analysis via Inter-objective Trade-offs}
\begin{figure}[ht]
    \centering
    \includegraphics[width=1\linewidth] {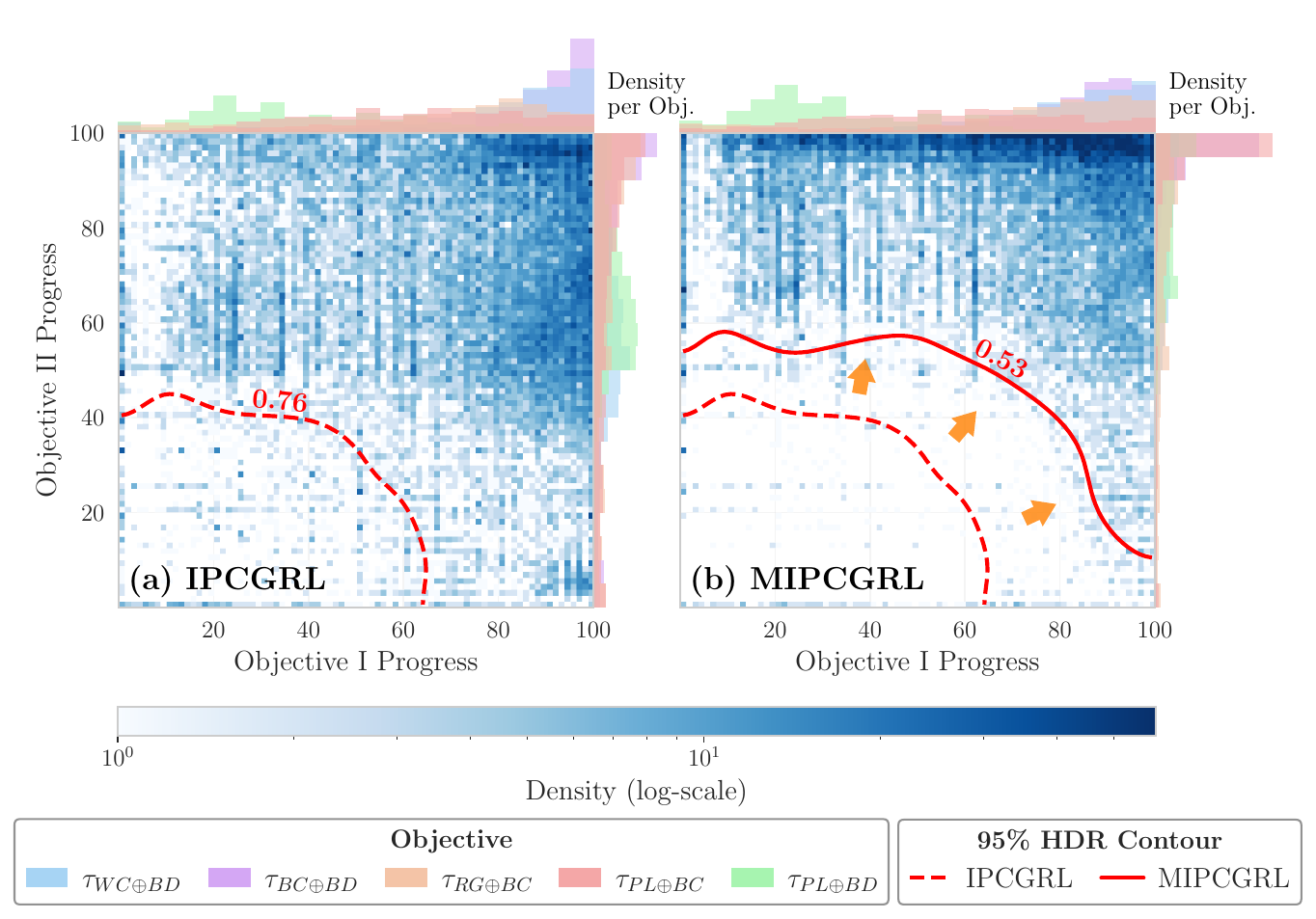}
    \vspace{-0.8cm}
    \caption{
    Density for multi-objective progress under IPCGRL (a) and MIPCGRL (b).
    Heatmaps show the estimated 2D density (log scale) with the density per objective.
    Red contours denote the 95\% HDR.
    Identical Gaussian smoothing ($\sigma=5$) is applied to both methods for stable contour extraction.
    }
    \label{capability}
    \vspace{-8pt}
\end{figure}

In this section, we examine whether progress across distinct objectives is intrinsically conflicting, thereby inducing structural trade-offs in multi-objective optimization.
To capture the interactions at the distribution level, we analyze the joint distribution of per-objective progress over all multi-objective samples using the highest-density region (HDR) \cite{hyndman1996computing}.
We use a 95\% HDR to emphasize typical outcomes and reduce sensitivity to rare extremes, facilitating stable cross-method comparisons.
It makes it possible to compare methods in terms of both trade-off geometry (via the contour shape) and outcome concentration (via the enclosed area).

As shown in Fig.~\ref{capability}, MIPCGRL shifts the joint-progress distribution toward the upper-right region, indicating outcomes that are closer to jointly optimal progress than those obtained by IPCGRL.
Under IPCGRL, the red dotted contour displays a pronounced bend, suggesting that inter-objective conflicts arise early and rapidly restrict simultaneous improvement.
In contrast, under MIPCGRL, the red contour is less sharply curved and its turning point occurs later, which is consistent with reduced inter-objective interference and an expanded regime where multiple objectives can improve concurrently.
This qualitative shift is corroborated by the HDR ratio in the joint-progress space, which decreases from 0.76 (IPCGRL) to 0.53 (MIPCGRL)—a 30.2\% reduction—indicating a substantially tighter concentration of typical joint-progress outcomes (i.e., greater consistency) rather than merely a change in pointwise averages.
Taken together, these observations are indicative of attenuated negative interactions across objectives and a larger effective region of simultaneous multi-objective improvement under MIPCGRL. 
Nevertheless, residual conflicts persist in certain compositions (e.g., $\tau_{\text{PL} \oplus \text{BD}}$) even under MIPCGRL, consistent with reward sparsity and objective-difficulty imbalance in multi-objective RL.
In particular, disparities in the density and magnitude of objective-specific reward signals can bias learning updates toward comparatively easier objectives, delaying or stagnating progress on harder objectives.

\subsection{Architectural Ablation Study}

To evaluate the contribution of individual components in MIPCGRL, $\text{MIPCGRL}_{\text{w/o REG, w/o CLS}}$, which removes the multi-head regression (REG) and the objective classifier (CLS), respectively.
In addition, we include Cont (goal) as a scalar-conditioned baseline to disentangle the effect of language-based objective decomposition from scalar-conditioned control. 
By comparing these variants with the full MIPCGRL model and the IPCGRL baseline, we analyze the role of objective-aware regression and instruction-level objective selection in instruction-conditioned multi-objective generation. The results of this ablation study are summarized in Fig. \ref{fig:result graph}.

Removing the objective classifier ($\text{MIPCGRL}_{\text{w/o CLS}}$) shows that most of the performance gain is retained by the regression module, particularly in the multi-objective setting (Fig.~\ref{fig:result graph} (b)).
In this case, a composed instruction is treated as a single target, and $\text{MIPCGRL}_{\text{w/o CLS}}$ remains largely comparable to the full model.
In contrast, in the single-objective setting (Fig.~\ref{fig:result graph} (a)), the performance degrades consistently without CLS (e.g., $\tau_{\text{WC}+\text{BD}}$, $\tau_{\text{BC}+\text{BD}}$, $\tau_{\text{PL}+\text{BC}}$, $\tau_{\text{PL}+\text{BD}}$)
This suggests that, when different objective types are optimized within an RL policy, explicitly identifying the objective associated with each episode helps reduce cross-objective interference.
The largest drop on $\tau_{\text{WC}+\text{BD}}$  further supports this interpretation, as it combines scalar-oriented and direction-oriented objectives.
Such heterogeneity can increase inter-objective interference and hinder disentanglement.
Overall, REG provides the primary objective-specific representational capacity, while CLS serves as a complementary inductive bias that helps separate heterogeneous episode-level objectives.

\section{Conclusion and Future Work}
We propose MIPCGRL, a multi-objective instruction representation learning framework for language-instructed PCGRL.
MIPCGRL is designed to improve controllable generation under compositional multi-objective instructions by learning an objective-sensitive instruction encoder that disentangles objective information.
Concretely, it jointly trains (i) a multi-label objective classifier to infer which objectives are active and (ii) a multi-head fitness regression module to encode objective-wise condition intensity, while applying probabilistic masking to emphasize objective-relevant features.
This decomposition separates objective identity from objective strength, yielding more robust and interpretable conditioning signals for multi-objective policy learning.

Empirically, MIPCGRL substantially improves performance in the multi-objective setting, achieving a 13.2\% gain over IPCGRL \cite{ipcgrl}, and exhibits more stable optimization and generalization under multi-objective instructions. 
Additionally, MIPCGRL retains strong behavior in the single-objective setting, with comparable performance to IPCGRL—indicating that the proposed multi-objective representation does not sacrifice single-objective controllability.
Moreover, when scaling to more complex compositions involving three objectives, MIPCGRL continues to display coherent trends consistent with those observed in simpler constituent compositions, suggesting that the learned representation integrates additional constraints without collapsing objective-wise competency. 
Overall, these results demonstrate that MIPCGRL effectively mitigates negative transfer and enhances stability as instruction complexity increases, providing a practical foundation for controllable multi-objective content generation.
Nevertheless, residual conflicts persist for certain compositions, plausibly due to the reward magnitude and frequency. Future work will focus on developing adaptive reward weighting mechanisms that dynamically adjust gradient normalization~\cite{gradnorm} based on the agent's learning state.

\section{Acknowledgments}
This work was supported by Institute of Information \& communications Technology Planning \& Evaluation (IITP) grant funded by the Korea government (MSIT) (No.2019-0-01842, Artificial Intelligence Graduate School Program (GIST)). 
This research was supported by the ‘Project for science and technology opens the future of the region’ program, funded by the Ministry of Science and ICT(MSIT) and Gwangju Metropolitan City, Republic of Korea in 2026 (Project Name:Convergence culture virtual studio for realizing artificial intelligence based metaverse).

\bibliographystyle{IEEEtran}
\bibliography{references}

\end{document}